\documentclass{article}

    \PassOptionsToPackage{numbers, compress}{natbib}

    \usepackage[final]{neurips_2022}

\usepackage[utf8]{inputenc} 
\usepackage[T1]{fontenc}    
\usepackage{hyperref}       
\usepackage{url}            
\usepackage{booktabs}       
\usepackage{amsfonts}       
\usepackage{nicefrac}       
\usepackage{microtype}      
\usepackage{xcolor}         
\usepackage{graphicx}
\usepackage{footmisc}
\usepackage{booktabs}
\usepackage{tablefootnote}
\usepackage{amsmath}
\usepackage{caption}
\usepackage{graphicx}
\usepackage{placeins}
\usepackage{pgfplots}
\usepackage{pgfplotstable}
\usepackage{tikz}
\usepackage{adjustbox}
\usepackage{amsmath}
\usepackage{float}
\usepackage{footnote}
\usepackage{etoolbox}
\usepackage{array, booktabs, makecell}
\usepackage{subcaption}
\usepackage{nameref}
\usepackage{hyperref}

\makeatletter
\newcommand*{\centerfloat}{%
  \parindent \z@
  \leftskip \z@ \@plus 1fil \@minus \textwidth
  \rightskip\leftskip
  \parfillskip \z@skip}
\makeatother

\title{QuaLA-MiniLM: a Quantized Length Adaptive MiniLM}

\author{%
 Shira Guskin \\
  Intel Labs \\
  \texttt{shira.guskin@intel.com}
  \And
  Moshe Wasserblat \\
  Intel Labs \\
  \texttt{moshe.wasserblat@intel.com}
  \And
  Chang Wang \\
  Intel \\
  \texttt{chang1.wang@intel.com}
  \And
  Haihao Shen \\
  Intel \\
  \texttt{haihao.shen@intel.com}
}

\begin{document}

\maketitle

\begin{abstract}
Limited computational budgets often prevent transformers from being used in production and from having their high accuracy utilized. A knowledge distillation approach addresses the computational efficiency by self-distilling BERT into a smaller transformer representation having fewer layers and smaller internal embedding. However, the performance of these models drops as we reduce the number of layers, notably in advanced NLP tasks such as span question answering. In addition, a separate model must be trained for each inference scenario with its distinct computational budget. Dynamic-TinyBERT tackles both limitations by partially implementing the Length Adaptive Transformer (LAT) technique onto TinyBERT, achieving x3 speedup over BERT-base with minimal accuracy loss. In this work, we expand the Dynamic-TinyBERT approach to generate a much more highly efficient model. We use MiniLM distillation jointly with the LAT method, and we further enhance the efficiency by applying low-bit quantization. Our quantized length-adaptive MiniLM model (QuaLA-MiniLM) is trained only once, dynamically fits any inference scenario, and achieves an accuracy-efficiency trade-off superior to any other efficient approaches per any computational budget on the SQuAD1.1 dataset (up to x8.8 speedup with <1\% accuracy loss). The code to reproduce this work is publicly available on Github\footnote{\url{https://github.com/intel/intel-extension-for-transformers/tree/main/examples/huggingface/pytorch/question-answering/dynamic}}.

\end{abstract}

\section{Introduction}
\label{sec:intro}
In recent years, increasingly large transformer-based models such as BERT~\cite{Devlin2019BERTPO}, RoBERTa~\cite{Liu2019RoBERTaAR} and GPT-3~\cite{Brown2020LanguageMA} have demonstrated remarkable state-of-the-art (SoTA) performance in many Natural Language Processing (NLP) and Computer Vision (CV) tasks and have become the de-facto standard. 
However, those models are extremely inefficient; they require massive computational resources and large amounts of data as basic requirements for training and deploying. This severely hinders the scalability and deployment of AI-based systems across the industry.

One highly effective method for improving efficiency is knowledge distillation~\citep{Ba2014DoDN,Hinton2015DistillingTK}, in which the knowledge of a large model defined as the teacher is transferred into a smaller more efficient model defined as the student. TinyBERT~\cite{Jiao2020TinyBERTDB} and MiniLM~\cite{Wang2020MiniLMDS,Wang2021MiniLMv2MS} stand out with their superior accuracy-speed-size tradeoff, both introducing a novel distillation method specially designed for transformers. In TinyBERT, the knowledge residing in the hidden states and attention matrices of the teacher is transferred to the student, and in MiniLM the student deeply mimics the teacher's self-attention module by transferring the multi-head self-attention relations as computed by a scaled dot-product of pairs of queries, keys and values. Both methods consist of a two-stage learning framework which first captures the general-domain knowledge of the teacher and then captures the task-specific knowledge of the teacher using fine-tuning. MiniLM achieves higher accuracy than TinyBERT for most downstream tasks, with the same number of parameters, and is much simpler to fine-tune since it is directly fine-tuned on the downstream task, differently from TinyBERT which is fine-tuned using a task-specific distillation from a fine-tuned teacher for a large number of epochs. In addition, TinyBERT fine-tuning requires a preliminary data-augmentation stage.

Knowledge distillation has shown promising results for reducing the number of parameters, with, however, several caveats: First, a drop in accuracy and a still limited speed-up/latency gain, specifically in challenging NLP tasks such as QA; for example, DistilBERT~\cite{Sanh2019DistilBERTAD} produces a x1.7 speed-up albeit with a 3\% accuracy drop on SQuAD1.1~\cite{Rajpurkar2016SQuAD1Q}. Secondly, in many cases the target computational budget (HW type, memory size, latency constraints, etc.) is not given at the time of training. This implies that a separate student model must be trained for each applicable inference scenario and its distinct computational budget.

Recent studies have attempted to address these concerns by proposing dynamic transformers. Length-Adaptive Transformer (LAT)~\cite{kim-cho-2021-length} introduced the Drop-and-Restore method, which reduces the sequence length at each layer, finally bringing back the dropped tokens in the last layer to allow for a wide range of NLP tasks despite the dropping of tokens. In addition, LAT proposed a one-shot training method that enables the model to be used for any computational budget during inference time.

Dynamic-TinyBERT~\cite{Guskin2021DynamicTinyBERTBT} utilizes TinyBERT~\cite{Jiao2020TinyBERTDB} distillation and some LAT techniques (Drop-and-Restore inference and evolutionary-search) to train an efficient model that can be used for a wide range of NLP tasks with optimal performance per any computational budget. Dynamic-TinyBERT achieves up to x3 speedup with <1\% accuracy loss vs. BERT-base on the SQuAD1.1 benchmark.

In this work we expand Dynamic-TinyBERT to generate a much more highly efficient model. First, we use a much smaller MiniLM model which was distilled from a RoBERTa-Large teacher rather than BERT-base. Second, we apply the LAT method to make the model length-adaptive, and finally we further enhance the model's efficiency by applying 8-bit quantization~\cite{Zafrir2019Q8BERTQ8}. The resultant QuaLA-MiniLM (Quantized Length-Adaptive MiniLM) model outperforms BERT-base with only 30\% of parameters, and demonstrates an accuracy-speedup tradeoff that is superior to any other efficiency approach (up to x8.8 speedup with <1\% accuracy loss) on the challenging SQuAD1.1 benchmark. Following the concept presented by LAT, it provides a wide range of accuracy-efficiency tradeoff points while alleviating the need to retrain it for each point along the accuracy-efficiency curve.

\begin{figure}[t]
    \centering
    \includegraphics[width=13cm, height=3.6cm]{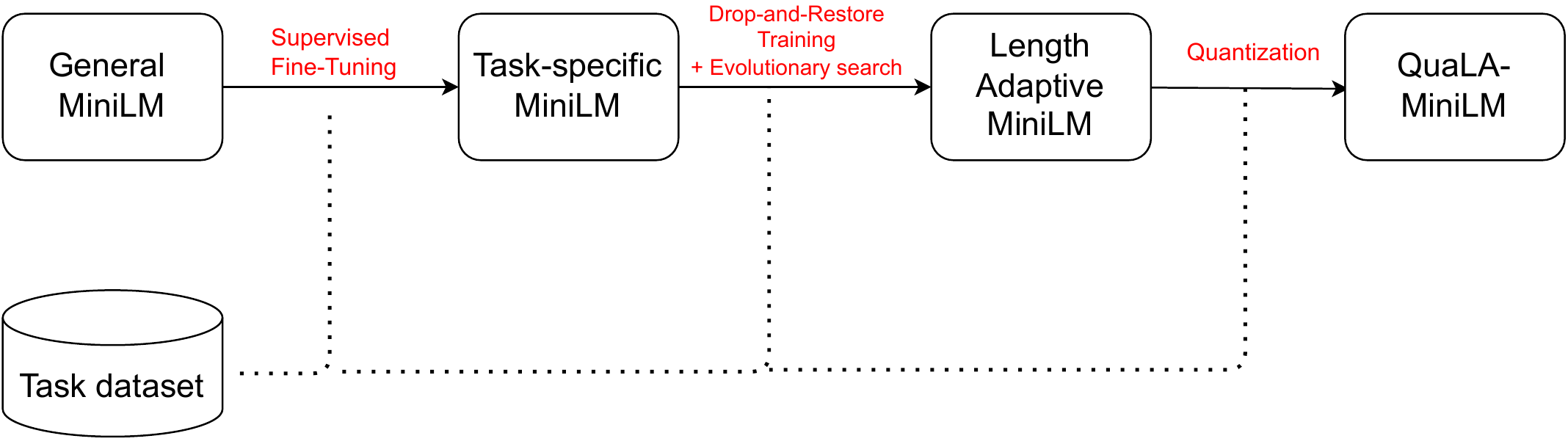}
    \caption{QuaLA-MiniLM training process. To run the model with the best accuracy-efficiency tradeoff per a specific computational budget, we set the length configuration to the best setting found by an evolutionary search to match our computational constraint.}
    \label{fig:quala_minilm_training_fig}
\end{figure}

\section{Method}

QuaLA-MiniLM is generated by applying the following optimization techniques on top of each other: MiniLM distillation, Length Adaptive Transformer, and Quantization. Figure~\ref{fig:quala_minilm_training_fig} demonstrates the training pipeline.

\subsection{MiniLM Distillation}
\label{sec:distillation}
MiniLM uses deep self-attention distillation to generate a task-agnostic small Language Model (LM). The student model learns to deeply mimic the multi-head self-attention relations -- which are obtained by a scaled dot-product of pairs of queries, keys and values of multiple relation heads of a single teacher's layer. Training a task-specific MiniLM requires two steps: first, training a task-agnostic MiniLM using multi-head self-attention distillation on a general-domain's data; second, fine-tuning it to a specific downstream task using standard supervised fine-tuning on the task dataset.

\begin{table}[t]
\centering
\caption{Inference performance on the SQuAD1.1 evaluation dataset. For all the length-adaptive (LA) models we show the performance both of running the model without token-dropping, and of running the model in a token-dropping configuration according to the optimal length configuration found to meet our accuracy constraint.}

\label{tab:models-performance-table}
\resizebox{0.9\columnwidth}{!}{%
\begin{tabular}{@{}lllllll@{}}

\toprule
 Model  & Model size (Mb) &  Tokens per layer & Accuracy (F1) &  Latency (ms) & FLOPs & Speedup        \\ \bottomrule
BERT-base   & 415.4723 & (384,384,384,384,384,384) & 88.5831 & 56.5679 & 3.53E+10 & 1x          \\  \bottomrule
TinyBERT-ours  & 253.2077 & (384,384,384,384,384,384) & 88.3959  & 32.4038 & 1.77E+10 & 1.74x    \\
QuaTinyBERT-ours  & 132.0665 & (384,384,384,384,384,384) & 87.6755  & 15.5850 & 1.77E+10 & 3.63x    \\ \bottomrule
MiniLMv2-ours   & 115.0473 & (384,384,384,384,384,384) & 88.7016  & 18.2312 & 4.76E+09 & 3.10x  \\
QuaMiniLMv2-ours   & 84.8602 & (384,384,384,384,384,384) & 88.5463  & 9.1466 & 4.76E+09 & 6.18x  \\ \bottomrule
LA-MiniLM   & 115.0473 & (384,384,384,384,384,384) & 89.2811  & 16.9900 & 4.76E+09  & 3.33x         \\ 
LA-MiniLM & 115.0473 & (269, 253, 252, 202, 104, 34) & 87.7637 & 11.4428 & 2.49E+09  & 4.94x          \\ \bottomrule
QuaLA-MiniLM & 84.8596 &(384,384,384,384,384,384) &  88.8593  & 7.4443 & 4.76E+09 & 7.6x       \\  
QuaLA-MiniLM & 84.8596 & (315,251,242,159,142,33) & 87.6828  & 6.4146 & 2.547E+09 & \textbf{8.8x}          \\ \bottomrule
\end{tabular}

}

\end{table}

\subsection{LAT (Length Adaptive Transformer)}
\label{sec:lat}
After fine-tuning a transformer to a specific task, the model is trained with LengthDrop and LayerDrop in a process called Drop-and-Restore, in which tokens are dropped at a random rate at each layer, and are brought back in the last hidden layer to enable a wider range of tasks and to make the model robust to the choice of length configuration at inference time. A length configuration
is a sequence of retention parameters (l\textsubscript{1}, · · · l\textsubscript{L}) each of which corresponds to the number of word vectors that are kept at each layer. Drop-and-Restore training is done using inplace distillation and sandwich rule methods, as follows: the full model without LengthDrop is fine-tuned for the downstream task as usual by minimizing the supervised loss function, while simultaneously, randomly-sampled sub-models with length reduction (sandwiches) learn to mimic the predictions of the full model using knowledge distillation~\cite{Hinton2015DistillingTK}.
As proposed by LAT, we run a multi-objective evolutionary search~\cite{Cai2020OnceFA,Wang2020HATHT} to optimize the accuracy-efficiency trade-off per each computational budget without additional training. The algorithm finds the optimal length configurations per possible computational constraints by generating an initial population of length configurations, and evolving the population at each iteration by mutation and crossover to consist only of configurations that lie on a newly updated accuracy-efficiency Pareto frontier. This process repeats for many iterations until best tradeoff is found.

\subsection{Quantization}
\label{sec:quantization}
We use the traditional post-training quantization~\cite{jacob2018quantization,wang2019haq} to quantize the model. Quantization for neural network is the process of approximating used floating-point numbers \textit{r} by low bit width numbers \textit{q} by the following mapping: \textit{r=S(q-Z)} where \textit{S, Z} represents the scale-factor and the zero-point values. Post-training quantization requires calibration of samples from representative datasets and collection of tensor statistics such as min and max values to determine the scale-factor and zero point values. Typically, we expect an INT8 8-bit model to gain more instruction throughput over the FP32 model (e.g., x4 for Intel DL Boost) and to gain about x4 lower memory bandwidth over the FP32 model, and therefore deliver higher inference efficiency.

\section{Experiments Setup}

We first fine-tune a pre-trained general-MiniLM for 5 epochs on the SQuAD1.1 dataset. We use the publicly available pre-trained MiniLMv2 with 6-layers and a hidden size of 384 distilled from RoBERTa-large\footnote{\url{https://github.com/microsoft/unilm/tree/master/minilm}}. We train our task-specific MiniLM to be length adaptive by running another 10 epochs of fine-tuning with Drop-and-Restore. After training, we run an evolutionary search to identify the length configurations that maximize the performance per any computational budget, namely those that achieve the best tradeoff between high accuracy and low number of floating operations (F1 vs. FLOPs). We leverage an open-source quantization tool~\cite{ipex} to obtain a QuaLA-MiniLM. See appendix section for a full description of system configurations. Results may vary.

\section{Results}
\label{sec:results}

\begin{figure}[t]
\centerfloat
\resizebox{0.7\textwidth}{!}{

\begin{tikzpicture}
\definecolor{s1}{RGB}{197, 90, 17}
\definecolor{clr_orange}{RGB}{255, 127, 0}
\definecolor{clr_green}{RGB}{31, 182, 83}
\definecolor{clr_purple}{RGB}{182, 83, 204}
\definecolor{clr_fuchsia}{RGB}{145, 92, 130}
\begin{axis}[
    width=\linewidth,
    xlabel={FLOPs ratio [BERT-base FLOPs / model FLOPs]},
    ylabel={F1},
    xmin=1, xmax=16,
    ymax=90.5,
    ymajorgrids=false,
    legend columns=1,
    legend pos=north east,
    legend style={at={(0.7,0.99)},anchor=north west},
    xtick pos=left, ytick pos=left,
    xtick={1, 2,3,4, 5, 6,7,8,9,10,11,12,13,14,15,16},
    ytick={84, 84.5, 85,85.5,86,86.5,87,88, 88.5, 89, 89.5 ,90, 90.5, 91},
    extra y ticks = {87.5},
    extra y tick style = {brown},
    yticklabels={84,84.5,85,85.5,86,86.5,87,88, {BERT 88.5}, 89, 89.5, 90, 90.5, 91},
    xticklabels={1, 2,3,4,5, 6,7,8,9,10,11,12,13,14,15,16},
    typeset ticklabels with strut,
    enlarge x limits=false,
    scale only axis]
    \addlegendimage{blue}
    \addlegendimage{green}
    \addlegendimage{yellow}
    \addplot[green, mark=*] table [x=speedup, y=f1, col sep=comma] {evo_minilm_ld.csv};
    \addplot[blue, mark=*] table [x=speedup, y=f1, col sep=comma] {quala_minilm.csv};
    \addplot[yellow, mark=*] table [x=speedup, y=f1, col sep=comma] {DynamicTinyBERT.csv};
    \addplot[black ,mark=*, mark size=3pt, nodes near coords={TinyBERT}, every node near coord/.append style={xshift=20pt,yshift=3pt}] coordinates{(2, 87.5)}; 
    \addplot[black ,mark=*, mark size=3pt, nodes near coords={DistilBERT}, every node near coord/.append style={xshift=25pt,yshift=6pt}] coordinates{(1.3, 85.8)}; 
     \addplot[black ,mark=*, mark size=3pt, nodes near coords={MiniLM}, every node near coord/.append style={xshift=-25pt,yshift=-6pt}] coordinates{(7.42, 88.7)}; 
    \addplot[black ,mark=*, mark size=3pt] coordinates{(1, 88.5)};
    \addplot[mark=none, black] coordinates {(1,88.5) (16,88.5)}; 
    \addplot[mark=none, brown] coordinates {(1,87.5) (16,87.5)}; 

    \addlegendentry{QuaLA-MiniLM}
    \addlegendentry{LA-MiniLM}
    \addlegendentry{Dynamic-TinyBERT}
\end{axis}

\end{tikzpicture}
}

\hfill

\vspace{2mm}
\caption{We evaluate F1 vs. FLOP-ratio of LA-MiniLM and QuaLA-MiniLM models configured with each of the optimal length configurations found by the evolutionary search. Each length configuration determines a single performance result of the model. FLOP-ratio is computed by dividing the number of FLOPs used to run BERT-base by the number of FLOPs used to run the model with a specific length-configuration. Evaluation is done on the SQuAD1.1 evaluation set.}\label{fig:graph}


\end{figure}
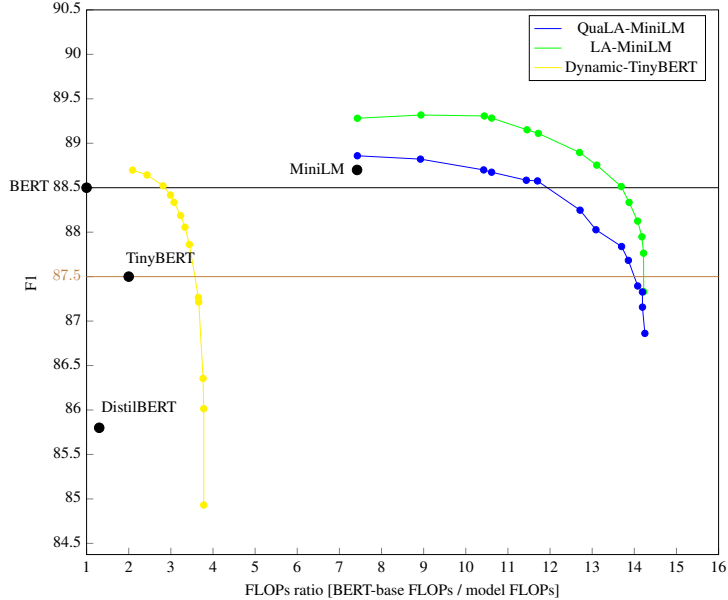

\subsection{Accuracy-efficiency trade-off}

Figure~\ref{fig:graph} shows the significant boost to inference performance achieved by our approach. Both LA-MiniLM and QuaLA-MiniLM outperforms BERT-base, Dynamic-TinyBERT, TinyBERT, and DistilBERT models in terms of both accuracy and efficiency, with up to x14 reduction of FLOPs and less than 1\% accuracy drop vs. BERT-base. QuaLA-MiniLM exhibits lower accuracy than LA-MiniLM with the same FLOPs count (since compression of weights from 32-bit to 8-bit does not affect the number of floating/int operations), but runs x2 faster due to its x4 instruction throughput gain (see the next section for a speedup gain analysis).

\subsection{Inference speedup}

Inference performance in terms of model size and latency is presented in Table~\ref{tab:models-performance-table}. The QuaLA-MiniLM model is x5 smaller than BERT-base, x3 smaller than TinyBERT, and runs about x8.8 faster than BERT-base, x5 than TinyBERT and x2 than quantized-TinyBERT, while sacrificing less than 1\% accuracy vs. BERT-base. LA-MiniLM configured with the optimal length configuration achieves x1.5 inference speedup over the LA-MiniLM model without token-dropping.

\section{Conclusions and future work}

We propose QuaLA-MiniLM, which leverages sequence-length reduction and low-bit representation techniques to further enhance MiniLM inference performance, enabling adaptive sequence-length sizes to accommodate different computational budget requirements with an optimal accuracy-efficiency tradeoff. Experiments on the SQuAD1.1 benchmark dataset demonstrate the effectiveness of our method compared with previous work on BERT compression. In future work we intend to explore how leveraging Sparsity can further boost MiniLM performance.

\bibliographystyle{abbrvnat}
\bibliography{references}

\appendix

\section{Appendix}

\subsection{Datasets}

In all our experiments we used the SQuAD1.1 dataset available at Huggingface's Datasets python library: https://github.com/huggingface/datasets.

\subsection{Hyper-parameters}

For all the training steps we used a maximum sequence length of 384. For the MiniLM fine-tuning we use a batch size of 8 and learning rate of 3e-5. We found these hyper-parameters to maximize our fine-tuned BERT-base performance, and due to time limitations we used the same settings for MiniLM. For both Drop-and-Restore and evolutionary search we use the same hyper-parameters used in LAT~\cite{kim-cho-2021-length}.

\subsection{System configurations}

 Inference performance was tested as of 09/12/2022 on a 2-node, 2x Intel® Xeon® Platinum 8280 Processor, 28 cores, HT On, Turbo ON, Total Memory 192 GB (12 slots/ 16GB/ 2934 MHz), BIOS: SE5C620.86B.02.01.0008.031920191559(0x5003006), CentOS Linux release 8.4.2105, gcc 9.3.0 compiler, Transformer-based Models, Deep Learning Framework: PyTorch 1.12 \url{https://download.pytorch.org/whl/cpu/torch-1.12.0%2Bcpu-cp38-cp38-linux_x86_64.whl}, IPEX 1.12 \url{http://intel-optimized-pytorch.s3.cn-north-1.amazonaws.com.cn/wheels/v1.12.300/intel_extension_for_pytorch-1.12.300%2Bcpu-cp38-cp38-linux_x86_64.whl}, BS=32, Public Data, 1 sockets, Datatype: FP32/INT8.
 For training the models we use a Titan-V GPU.

\end{document}